\documentclass[10pt,twocolumn,letterpaper]{article}

\usepackage{cvpr}              % To produce the CAMERA-READY version
\usepackage{graphicx}
\usepackage{amsmath}
\usepackage{amssymb}
\usepackage{booktabs}
\usepackage{makecell}

\usepackage[pagebackref,breaklinks,colorlinks]{hyperref}
\usepackage{multirow}

\usepackage{geometry}
\usepackage[capitalize]{cleveref}
\crefname{section}{Sec.}{Secs.}
\Crefname{section}{Section}{Sections}
\Crefname{table}{Table}{Tables}
\crefname{table}{Tab.}{Tabs.}

\def\cvprPaperID{*****} % *** Enter the CVPR Paper ID here
\def\confName{CVPR}
\def\confYear{2022}

\begin{document}

\title{The Second-place Solution for CVPR 2022 SoccerNet Tracking Challenge}

\author{
Fan Yang, Shigeyuki Odashima, Shoichi Masui, Shan Jiang\\
Fujitsu Research\\
{\tt\small {contact: fan.yang}@fujitsu.com}
}
\maketitle

%%%%%%%%% BODY TEXT
\section{Summarization}
This is our second-place solution for CVPR 2022 SoccerNet Tracking Challenge~\cite{giancola2022soccernet}. Our method mainly includes two steps: online short-term tracking using our Cascaded Buffer-IoU (C-BIoU) Tracker~\cite{CBIOU_2023_WACV}, and, offline long-term tracking using appearance feature and hierarchical clustering. At each step, online tracking yielded HOTA scores near $90$, and offline tracking further improved HOTA scores to around $93.2$.
%-------------------------------------------------------------------------

\section{Method Details}
\subsection{Online Tracking}

\begin{figure}[!h]
  \centering
  \includegraphics[width=\linewidth]{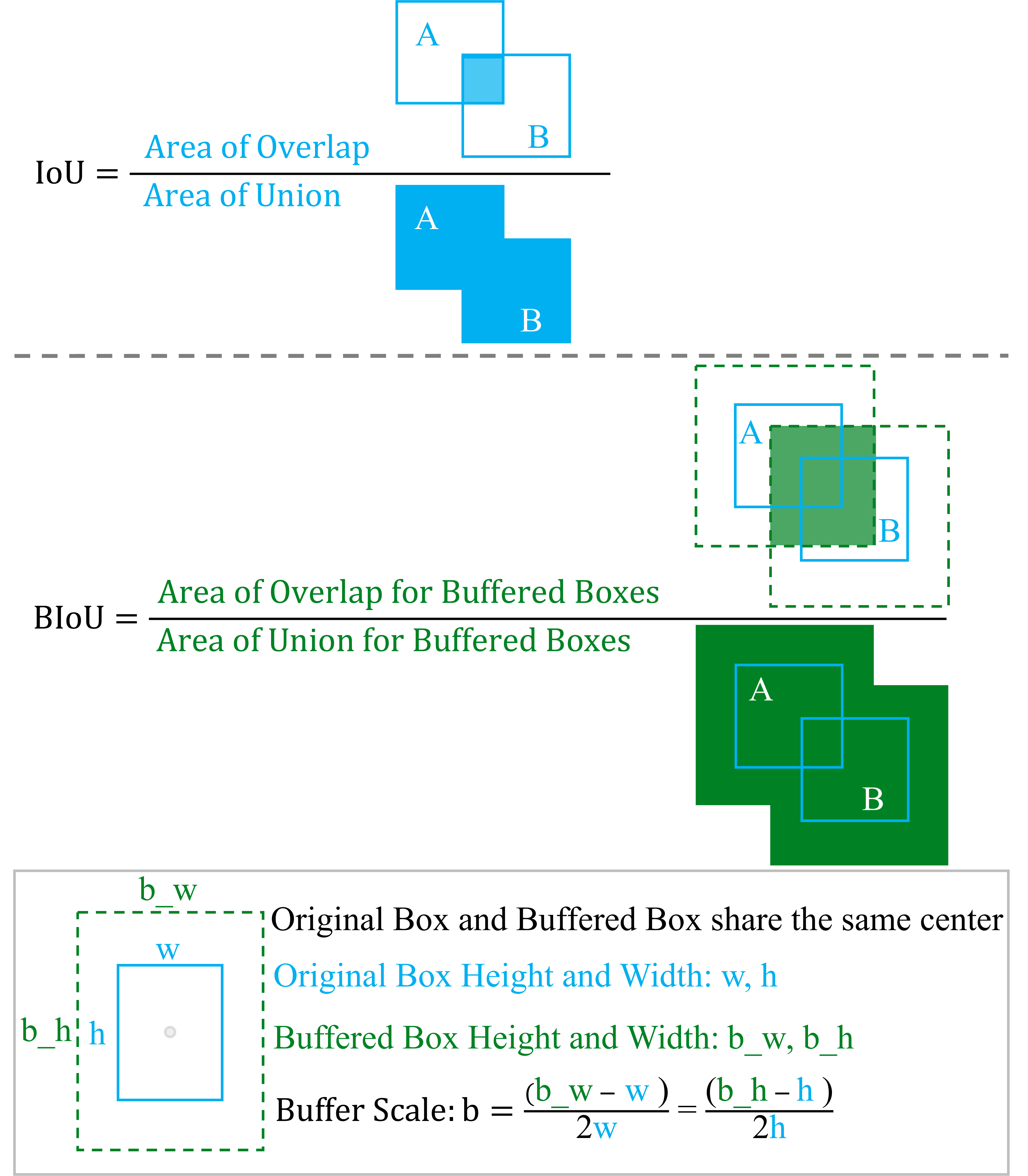}
  \captionsetup{font=small}
  \caption{Comparison between IoU and our BIoU.}
  \label{fig:biou}
\end{figure}

We generated short-term tracklets by solely using geometry features. In general, IoU is commonly used to calculate cross-frame geometric affinity between observations~\cite{IoUTracker2017,DeepSORT}. However, unlike conventional MOT datasets that focus on pedestrian tracking, SoccerNet tracking data may face a special challenge: when players' movements are fast and complicated, conventional IoU matching may ignore non-overlapping observations for the identical player and cause ID Switches. To address this challenge, we proposed a Buffer-IoU (BIoU). As Fig.~\ref{fig:biou} shows, BIoU simply adds buffers that are proportional to the original box size for calculating IoU scores. 

To some extent, replacing IoU with GIoU~\cite{giou2019} and DIoU~\cite{zheng2020distance} also alleviates the non-overlapping issue as our BIoU does, but we verified that our BIoU may generate better results under the same conditions.

In MOT, cascaded matching is a commonly used approach: matching the confident and easy samples first and matching the ambiguous and hard samples later. For instance, ByteTrack\cite{zhang2021bytetrack} matches the confident detections at the beginning and then is followed by matching low confident detections; DeepSORT matches recent tracks before older ones. We specifically designed a cascaded BIoU matching in our tracker. 
We first match detections to previous-frame tracks based on BIoU scores using a small buffer scale, and then match remaining detections and tracks with BIoU scores using a large buffer scale. To select suitable buffer scales for our CBIoU Tracker, we applied grid searching on the training set. When BIoU matching threshold is set to $0.01$, buffer scales (\ie, [0.7, 1.0]) yield the maximum HOTA score, which is selected for our model.

\begin{figure*}[!h]
  \centering
  \includegraphics[width=\linewidth]{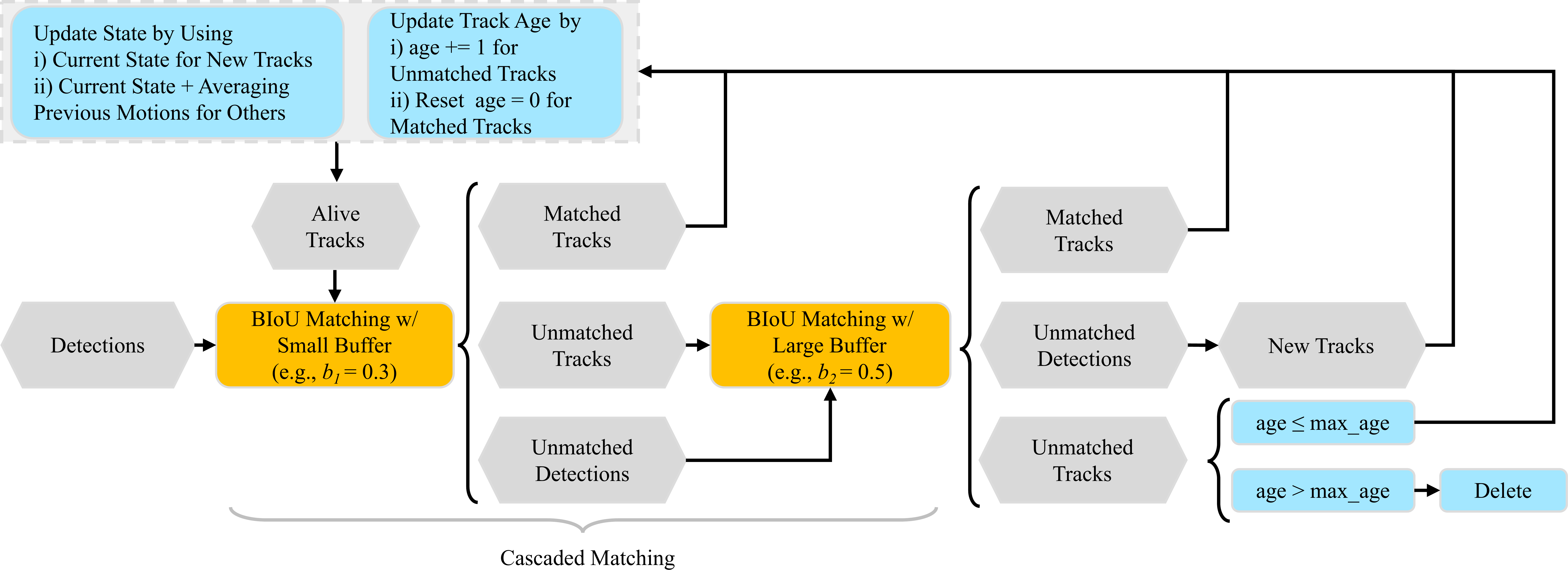}
  \captionsetup{font=small}
  \caption{The architecture of our Cascaded Buffer-IoU Tracker. A cascaded matching is performed by matching recent tracks first with a small buffer scale and matching the remaining tracks with a large buffer scale.}
  \label{fig:cbiout}
\end{figure*}

The architecture of our Cascaded BIoU Tracker (Fig.~\ref{fig:cbiout}) is similar to SORT~\cite{Bewley2016_sort}, but we changed the motion estimation method from Kalman Filter to simply averaging speeds of the previous two frames. We suppose that soccer players may have non-linear motions, which are difficult to be modeled by Kalman filter. Our motion estimation therefore only utilizes recent speeds to quickly respond to unpredictable motion change. Meanwhile, our BIoU enlarges the matching space and can compensate for inaccurate motion estimation.

\subsection{Offline Tracking}
Since the camera view cannot cover the whole soccer field, some long-term tracklets could be broken down by only referring to geometry features. To recover long-term tracklets, we employed offline tracking with appearance features. 

We utilized Strong ReID~\cite{Luo_2019_Strong_TMM} to obtain the appearance feature. To initialize the re-id model, we utilized training data from SoccerNet 2022 re-id challenge. Since the re-id training data may contain noise, we only used those pairs that have more than 5 instances for our training. Although training and validation sets of tracking data have GT for tracklets, one game event might be split into multiple videos so that the player ID is not unique. To include the tracking labels in our re-id training, we took a self-supervised learning method introduced by ReMOT/ReMOTS~\cite{yang2021remot, yang2020remots}. Referring to tracking ID, in each video, we still can construct triplets and only apply triplet loss to them. 

After the training, we generated appearance features for each short-term tracklets obtained from the previous step. Within a video, we formed a distance matrix $\mathcal{D}$ between short-term tracklets as
\begin{equation}
\label{eq:app}
\scalebox{0.85}{
\begin{math}
\begin{aligned}
\nonumber
\mathcal{D}_{k1,k2} = &\\
&\left\{\begin{matrix}
inf,  ~~~~~~if~ \Pi _{k1} \cap \Pi _{k2}\neq \varnothing \\ 
\frac{1}{N_{k1}N_{k2}}
\sum_{i \in \Pi _{k1}}
\sum_{j \in \Pi_{k2}}
\big(1-\frac{f^{k1}_{i}f^{k2}_{j}}{\left \| f^{k1}_{i}\right \| \left \| f^{k2}_{j}\right \|} \big),   otherwise
\end{matrix}\right. 
\end{aligned}
\end{math}}
\end{equation}
where for tracklets $T_{k1}$ and $T_{k2}$, $\mathcal{D}_{k1,k2}$ is their distance; $\Pi _{k1}$ and $\Pi _{k2}$ are their temporal ranges; $f^{k1}_{i}$ and $f^{k2}_{j}$ are their appearance features at frame $i$ and $j$, and $N_{k1}$ and $N_{k2}$ are the number of observations within the tracklets, respectively.

\begin{figure}[!h]
  \centering
  \includegraphics[width=\linewidth]{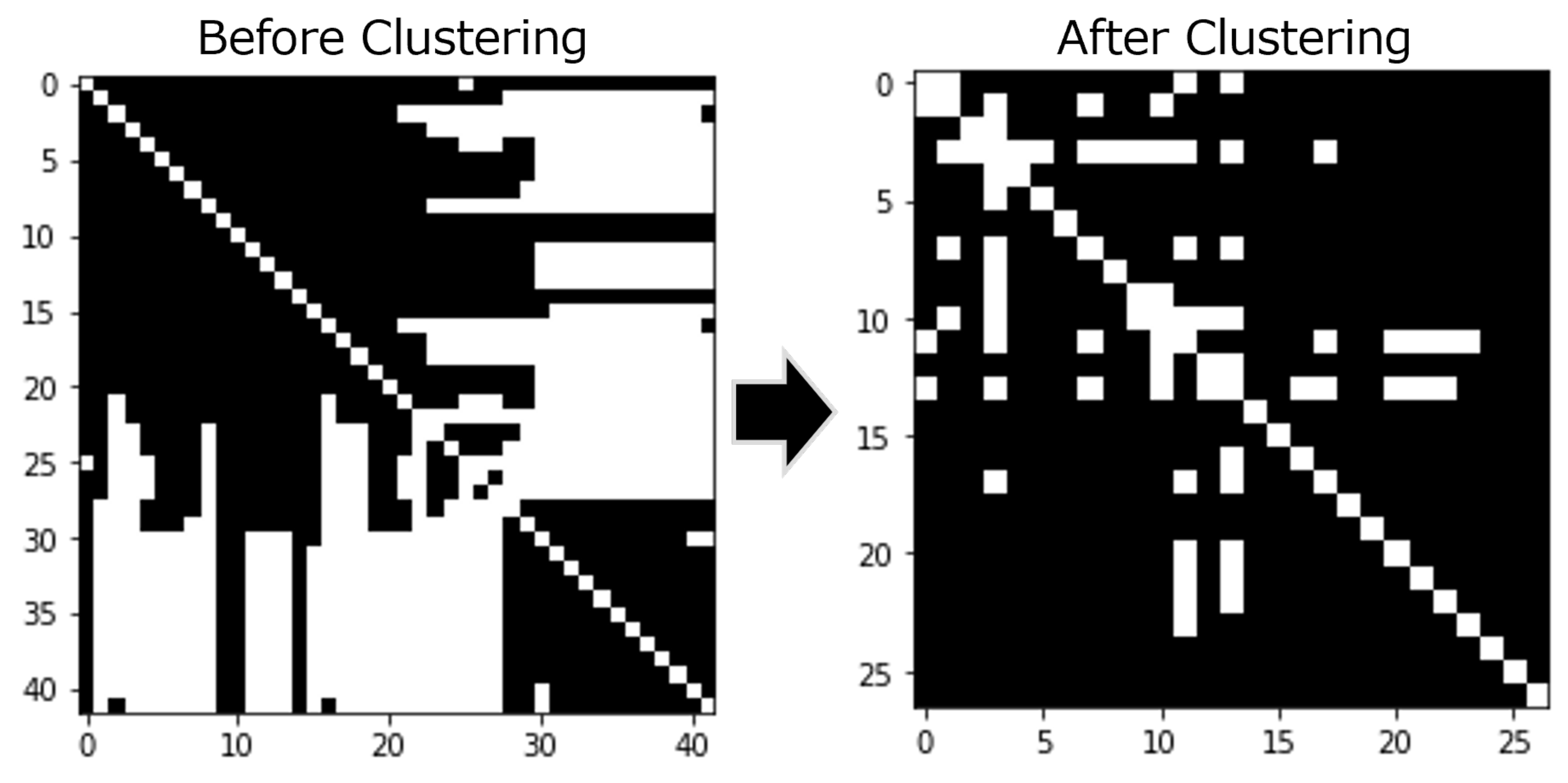}
  \captionsetup{font=small}
  \caption{The change of distance matrix $\mathcal{D}$ before and after clustering for one video example. After clustering, short-term tracklets are merged to long-term tracklets and the number of tracklets has decreased.}
  \label{fig:clustering}
\end{figure}

Based on $\mathcal{D}$, we applied hierarchical clustering (cutting threshold 0.15) to cluster short-term tracklets to long-term ones (see Fig.\ref{fig:clustering}).

\section{Results}

On the testing set, we compared a number of trackers by local evaluation. 
The results are listed in Table~\ref{table:soccer-net_test_oracle}. For the challenge set, we list our online and offline results in Table~\ref{table:soccer-net_challenge}. We applied $max\_age=60$ to C-BIoU to obtain its best performance. However, when we put C-BIoU and Offline Link together, we prefer C-BIoU generate shorter but cleaner tracklets, so we selected $max\_age=1$. To verify how the appearance feature affects the offline link results, we tested re-id models trained by Market~\cite{zheng2015scalable} and SoccerNet, respectively. Due to the domain gap, re-id model trained by SoccerNet obtained better result.

Due to the simplicity of our C-BIoU tracker, it can reach $680$ FPS on the testing set, by using an Intel Xeon Silver 4216 CPU. For our offline processing, extracting the appearance feature from images costs about $40$ mins on the testing set (on a single Nvidia GeForce 1050 GPU), and performing hierarchical clustering costs about $4$ mins.

\begin{table*}[t]
  \caption{Results on SoccerNet testing set with oracle detections (GIoU Tracker and DIoU Tracker are implemented by us). Although GIoU~\cite{giou2019} and DIoU~\cite{zheng2020distance} can incorporate non-overlapped boxes, applying them in MOT may not generate comparable results as our C-BIoU does. Integrating appearance features in our offline link can improve HOTA score since the uniforms are different between two soccer teams, and the referee also has a different uniform. }
  \centering
  \small
  \begin{tabular}{ l | ccccc}
  \toprule
  Tracker & HOTA$\uparrow$ & DetA$\uparrow$ & AssA$\uparrow$ & MOTA$\uparrow$ & IDF1$\uparrow$\\
  \midrule
  GIoU Tracker~\cite{giou2019} &79.8 & \textbf{99.7} & 63.8 & 97.8 & 73.4\\
  DIoU Tracker~\cite{zheng2020distance} & 84.8 & \textbf{99.7} & 71.2 & 99.2 & 79.9\\
  IoU Tracker~\cite{IoUTracker2017} & 81.9 & 99.4 & 67.5 & \textbf{99.8} & 75.7\\
  \hline
  C-BIoU (max\_age=60) & 89.2 & 99.4 & 80.0 & 99.4 & 86.1\\
  C-BIoU (max\_age=60) + Offline Link (train re-id by SoccerNet) & 91.4 & \textbf{99.7} & 83.7 & 99.6 & 89.4\\
  C-BIoU (max\_age=1) & 88.1 & 99.5 & 77.9 & 99.4 & 83.3\\
  C-BIoU (max\_age=1) + Offline Link (train re-id by Market) & 90.9 & 99.5 & 82.9 & 99.5 & 89.2\\
  C-BIoU (max\_age=1) +  Offline Link (train re-id by SoccerNet) & \textbf{92.8} & 99.5 & \textbf{86.4} & 99.5 & \textbf{91.3}\\
  \bottomrule
  \end{tabular}

  \label{table:soccer-net_test_oracle}
\end{table*}

\begin{table*}[h!]
  \caption{Results on SoccerNet challenging set with oracle detections.}
  \centering
  %\small
  \begin{tabular}{ l | ccccc}
  \toprule
  Tracker & HOTA$\uparrow$ & DetA$\uparrow$ & AssA$\uparrow$ & MOTA$\uparrow$ & IDF1$\uparrow$\\
  \midrule
  C-BIoU (max\_age=60) &  89.6 & 99.6 & 80.7 & 99.6 & 87.5\\
  C-BIoU (max\_age=1) + Offline Link (train re-id by SoccerNet)  &  \textbf{93.2} & \textbf{99.8} & \textbf{87.2} & \textbf{99.7} & \textbf{91.6}\\
  \bottomrule
  \end{tabular}
  \label{table:soccer-net_challenge}
\end{table*}

\section{Other Explorations}
Besides previous reports, we did other explorations on this challenging dataset. 

\noindent
\textbf{1. We explored removing camera motions for tracking}.
We calculated the homography matrix between two adjacent frames to remove the camera motion for tracking, but there was no change in tracking results. This suggested that the camera motion may have little effect on tracking when oracle detections are obtained.

\noindent
\textbf{2. We explored performing tracking on Bird's-Eye-View (BEV) soccer field}.
By following \cite{cioppa2021camera}, we calculated homography from 2D image to the BEV soccer field. We mapped players' locations to the BEV coordinate for tracking, however, we obtained worse results on the testing set, with HOTA score of $77.6$. We suppose that homography bias may impair the tracking performance.

\begin{figure}[!h]
  \centering
  \includegraphics[width=\linewidth]{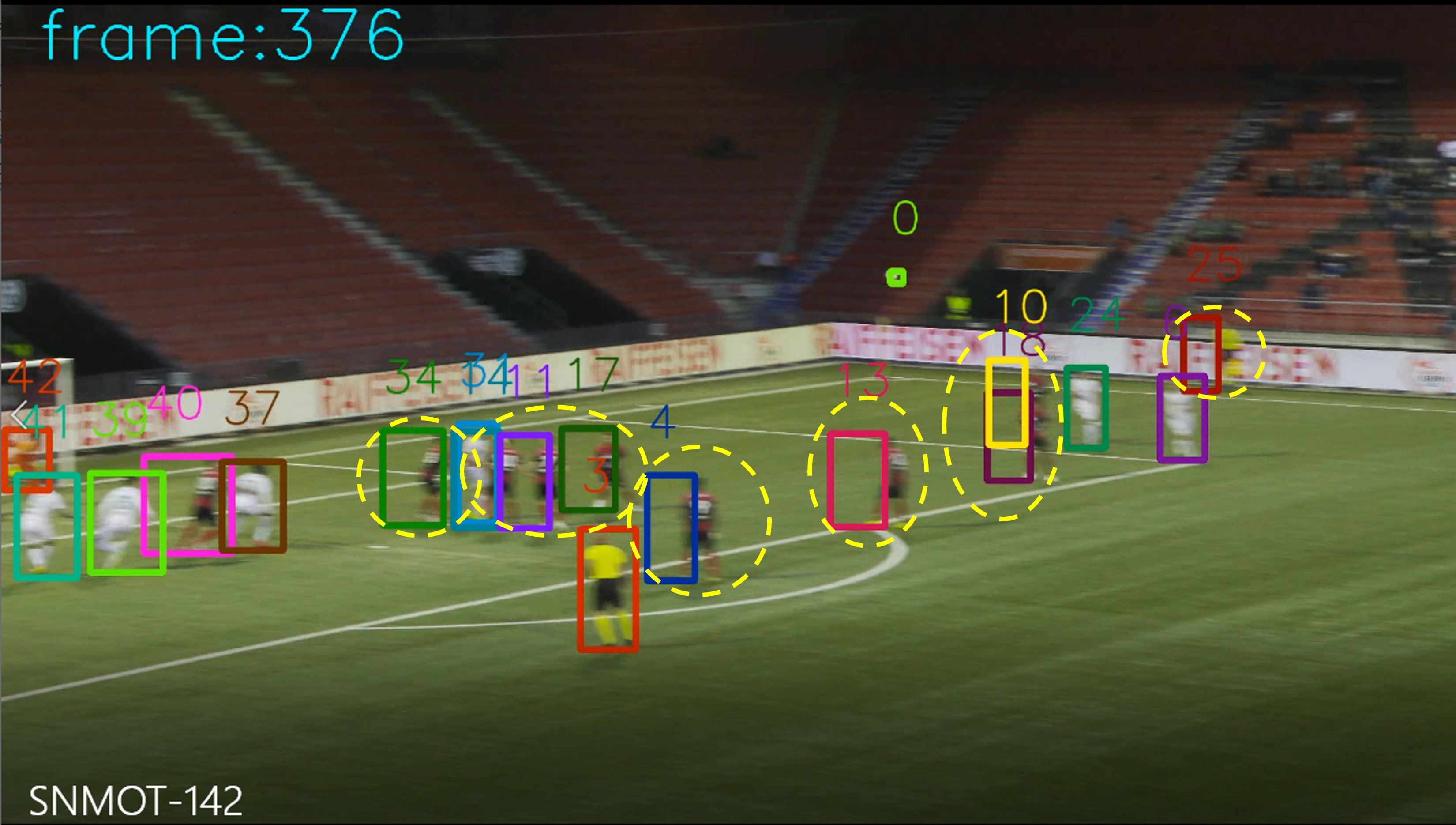}
  \captionsetup{font=small}
  \caption{An example of inaccurate GT detections in the testing set. Inaccurate bounding boxes are highlighted by yellow circles.}
  \label{fig:bad_annotations}
\end{figure}

\noindent
\textbf{3. We found some inaccurate detections}. By visualizing the official GT detections, we noticed that some bounding boxes are not accurate, which could impair the tracking result. We give an example in Fig.\ref{fig:bad_annotations}.

%%%%%%%%% REFERENCES
{\small
\bibliographystyle{ieee_fullname}
\bibliography{egbib}
}

\end{document}